%% file: main.tex
\documentclass[10pt,twocolumn,letterpaper]{article}

\usepackage{wacv}
\usepackage{times}
\usepackage{epsfig}
\usepackage{graphicx}
\usepackage{caption}
\usepackage{subcaption}
\usepackage{amsmath}
\usepackage{amssymb}

\def\sec#1{Sec.~\ref{#1}}
\def\fig#1{Fig.~\ref{#1}}

\usepackage{color}

\wacvfinalcopy 

\def\wacvPaperID{606} 

\ifwacvfinal\fi
\makeatletter
\newcommand*\titleheader[1]{\gdef\@titleheader{#1}}
\AtBeginDocument{%
  \let\st@red@title\@title
  \def\@title{%
    \bgroup\normalfont\small\centering\@titleheader\par\egroup
    \vskip0.5em\st@red@title}
}
\makeatother

\titleheader{Published as a conference paper at 2018 IEEE Winter Conf. on Applications of Computer Vision (WACV'2018)}
\title{Object-based reasoning in VQA}
\author{tkornuta}

\begin{document}


\author{Mikyas T. Desta \\
IBM Research, Almaden\\
{\tt\small mtdesta@us.ibm.com}
\and
Larry Chen\thanks{Work done during an internship at IBM Research, Almaden.}\\
University of Chicago\\
{\tt\small larrychen@uchicago.edu}
\and
Tomasz Kornuta\\
IBM Research, Almaden\\
{\tt\small tkornut@us.ibm.com}
}

\maketitle
\ifwacvfinal\thispagestyle{empty}\fi

\begin{abstract}
Visual Question Answering (VQA) is a novel problem domain where multi-modal inputs must be processed in order to solve the task given in the form of a natural language.
As the solutions inherently require to combine visual and natural language processing with abstract reasoning, the problem is considered as AI-complete.
Recent advances indicate that using high-level, abstract facts extracted from the inputs might facilitate reasoning.
Following that direction we decided to develop a solution combining state-of-the-art object detection and reasoning modules.
The results, achieved on the well-balanced CLEVR dataset, confirm the promises and show significant, few percent improvements of accuracy on the complex "counting" task.

\end{abstract}


\input{introduction}

\input{related_work}

\input{our_solution}

\input{results}

\section{Conclusions}
\label{sec:conclusions}
In this paper we have focused on reasoning using high-level  abstract facts.
For that purpose, instead of relying on features extracted from the images, we used facts in the form of encoded objects detected by the Faster R-CNN detector.
Those abstract facts were next passed to a reasoning module, developed with the aim of learning object-object relations.
The achieved overall accuracy is comparable with the current state-of-the-art results.
Analysis of the results unveiled that the proposed solution gives more stable results for different tasks, and, moreover, shows improvement in 2 out of 5 CLEVR tasks.
In particular, it gives significant, few percent improvement in the Counting task, which is currently considered as one of the most complex tasks in VQA and is gaining more and more attention, e.g. in~\cite{trott2017interpretable} the authors proposed a new dataset called HowMany-QA, devoted only to that specific problem.

The detailed analysis of the operation of components of our system has proven that OD is working properly, whereas difference in accuracy comes from the Relational Network module.
According to the results reported in~\cite{santoro2017simple} the RN module working in separation is supposed to give around 2\% better overall accuracy than the accuracy that we have managed to achieve.
We treat that as a general promise that RN operating on more abstract features should be able to achieve even better results.

As the used reasoning module is the bottle-neck of the system, in our future works we want to focus on this part of the system and experiment with other approaches. 
The most important research directions, aiming at improvement of our system and overcoming its limitations, are as follows.
The first limitation is the used aggregation of relations detected by MLPs -- as the number of objects in scenes might vary, utilization of recurrent neural net seems to be natural.
An interesting solution to that problem was recently proposed in~\cite{palm2017recurrent}, where the authors developed a novel neural-based message passing mechanism called Recurrent Relational Networks and have shown its superiority on Sudoku puzzles and the BaBi textual QA.
Second limitation of our system is that the reasoning performed in the Semantic Embedding module is strictly sequential, with hardcoded number of reasoning steps, whereas number of reasoning steps should in fact vary, depending on the task given. One of the possible solutions addressing that problem is to employ the Adaptive Computation Time (ACT) mechanism proposed in ~\cite{graves2016adaptive}.
Yet another promising direction is utilization of external memory (i.e. memory augmented neural networks such as Neural Turing Machine~\cite{graves2014neural}) for learning more complex reasoning schemes and memorization of the abstract, graph like representation of the observed scene before generating the final answer.
The latter is a natural extension for reasoning over explicit high-level representations of the contents of the image and shows very good results as reported e.g. in~\cite{wang2016vqa}.
Finally, we also want to improve our model by training it jointly on several tasks (e.g. word-level modeling, object detection and VQA) -- it was recently shown e.g. in \cite{kaiser2017one} that such a multi-task learning enables the model to develop unified representation and results in improved overall accuracy.

\section*{Acknowledgements}
We would like to thank Alexis Asseman for setting up and managing our hardware, enabling us to run our experiments smoothly, Ahmet S. Ozcan for his insights and proofread, T.S. Jayram, Vincent Albouy, Ben Lyo  and other members of our Machine Intelligence team in IBM Research, Almaden, for critical feedback and discussions.

{\small
\bibliographystyle{ieee}
\bibliography{sample}
}

\end{document}

%% file: introduction.tex
\section{Introduction}

Due to successful application of machine learning, and deep learning \cite{lecun2015deep} in particular, we have lately observed great progress in traditional Computer Vision (CV) tasks such as image classification \cite{krizhevsky2012imagenet}, object detection and image segmentation \cite{girshick2016region}. 
These advances have pushed researchers’ towards more complex tasks such as image caption generation \cite{karpathy2015deep} and VQA~\cite{antol2015vqa,malinowski2014towards}.
Both of those tasks combine Computer Vision with Natural Language Processing (NLP) and high-level reasoning, thus require utilization of multi-modal knowledge beyond a single sub-domain (such as CV).
In the case of caption generation, however, there may be several valid captions describing a given image. Thus, it is hard to define a quantitative evaluation metric to track the progress.
In contrast, in VQA the image and question pairs might come with the ground-truth answers, enabling to measure the system accuracy. 
Moreover, as there might be many questions referring to the same image, the system is in fact forced to learn various reasoning types depending on the type of question.
For those reasons VQA is considered to be AI complete~\cite{antol2015vqa} and renews the hope of building machines that could pass the Turing test in open domains~\cite{malinowski2014towards}.

During the last three years several complex VQA datasets have been introduced, however, many of the solutions show only marginal improvements over the strong baselines.
As there are many influencing factors it is often hard to tell which modules of the system are working properly and why. 
The CLEVR dataset~\cite{johnson2017clevr} was created with aim of pushing forward the progress on VQA in a more systematic manner, with the main focus put on the validation of reasoning skills.
The dataset delivers a well-balanced set of images and questions, where the information associated with each image is complete and exclusive. 
Along with the dataset the authors provided several, sometimes quite sophisticated baselines, and concluded that those models have not learned the semantics of spatial reasoning at all.

In this paper we decided to focus on that aspect of the problem, i.e. spatial reasoning.The hypothesis we would like to validate is whether the operation on high-level and abstract facts extracted from the image might improve the accuracy of the system (as suggested e.g. in~\cite{anderson2017bottom}).
For that reason, we developed a solution operating on object-relation-object triplets~\cite{dai2017detecting}, similar to relational network~\cite{santoro2017simple}, however, instead of using features extracted from the images, we rely on the information associated with the objects detected in the image. 
The achieved results are comparable with current state-of-the-art solutions in terms of overall accuracy and show significant improvement on the counting task.



%% file: related_work.tex
\section{Related work}
\label{sec:related_work}

Research on VQA has resulted in many interesting solutions.
Those, however, could not have been developed without the existence of proper datasets and metrics for evaluation and comparison of the results.

From the historical perspective, the first dataset designed as a benchmark for the VQA task was DAQUAR (DAtaset for QUestion Answering
on Real-world images)~\cite{malinowski2014multi}.
The dataset contains 1.5k RGB-D images of indoor scenes from NYU-Depth v2 dataset~\cite{silberman2012indoor}, with annotated semantic segmentations and question/answer pairs of two types: synthetic questions/answers generated automatically on the basis of NYU annotations and human question/answers collected from 5 annotators.
The limitations of DAQUAR (i.e. restriction of answers to a predefined set and strong bias in human annotations) resulted in releases of several other datasets, most prominently: COCO-QA ~\cite{ren2015image} (based on images from MS COCO dataset~\cite{lin2014microsoft}, where substantial effort was made in order to increase the scale of training data), VQA ~\cite{antol2015vqa,goyal2016making} (consisting of two sets, VQA-real with natural images, and VQA-abstract with cartoon images, and providing 17 additional (incorrect) candidate answers for each question) and Visual Genome~\cite{krishna2017visual} (currently the largest VQA dataset, with 1.7 million question/answer pairs and with images contents additionally described by structured annotations in the form of scene graphs).

The paper which introduced DAQUAR~\cite{malinowski2014multi} also laid the foundations for evaluation of the system accuracy, and in consequence, monitoring the overall progress in the field. 
The authors proposed two basic evaluation metrics. First, by simply measuring the accuracy with respect to the ground truth answer using string matching. Second, by using the Wu-Palmer Similarity (WUPS), enabling to evaluate the similarity between common objects by their distance in a taxonomy tree.
Other evaluation methods include the previously mentioned one in a multiple-choice settings~\cite{antol2015vqa} or the "fill in the blanks" approach proposed along with the Visual Madlibs dataset~\cite{yu2015visual}.
As the field started to mature, researchers discovered the importance of biasses in the images and questions, which triggered the release of more balanced datasets (such as VQA v2~\cite{goyal2016making} and CLEVR~\cite{johnson2017clevr}). It also led to the analysis of the importance of the common-sense knowledge required to solve a given task, which in turn resulted in efforts towards delivering ground truth containing rich description of the scene, facilitating both training and evaluation (e.g.~Visual Genome~\cite{krishna2017visual}). 

However, the most interesting achievements in the VQA field are the novel algorithms and neural architectures.
In~\cite{wu2017visual} the authors proposed to distinguish four categories: joint embedding approaches, attention mechanisms, compositional models
and models using external knowledge bases, which also summarizes the four main research directions in VQA.

The efforts in joint embedding focus on the methods for combining multi-modal representations.
As in VQA there are two distinct input modalities (image and text), which makes this problem similar to the problems found in other multi-modal domains.  For example the projection of user and item embeddings into the common representation space in neural recommender systems~\cite{he2017neural}.
Exemplary approaches developed for the VQA problem domain include e.g. Multimodal Compact Bilinear pooling (MCB)~\cite{fukui2016multimodal} method that performed joint embedding of visual and text features, or Relational Networks (RN)~\cite{santoro2017simple} where embedded question was concatenated with features extracted from pairs of image regions, enabling the system to reason about the relation between objects being present in those regions.

As some questions might require more than one reasoning step, a lot of researchers focused on attention mechanisms, which were initially introduced for neural translation ~\cite{bahdanau2014neural}, and later on adapted to the Question-Answering problem~\cite{weston2014memory}.
In~\cite{shih2016look} the authors introduced a simple attention model, where an embedded question was used for the generation of an attention mask (called region-question relevance), which was subsequently used for the calculation of a weighted vector of concatenated question and image features.
Yang et al. \cite{yang2016stacked} introduced Stacked Attention Networks, which are able to infer the answer iteratively.
This architecture includes two attention layers, enabling two reasoning steps (in~\cite{weston2014memory} called "hops"), each driven by a different attention mask over the image features.
The influence of number of hops was analysed in~\cite{xu2016ask}, where the authors introduced a spatial attention-based model called Spatial Memory Network. Here, the model uses each word embedding to capture a fine-grained alignment between the image and the question.
The results indicated the superiority of the two-hop model over the model with a single hop.
The described methods above focused only on visual attention driven by the question.
In contrast to those ideas, a major innovation was proposed in~\cite{lu2016hierarchical}, where Hierarchical Co-Attention (HieCoAtt) model processed image and question symmetrically, with the image guiding the attention over the question and vice versa.
In a quest to assess the quality of the attention models, in more recent works researchers have tried to validate whether the attention models learned by neural models focus on the same image regions as humans do~\cite{das2016human}.

\begin{figure*}[htbp]
\centerline{\includegraphics[width=\textwidth]{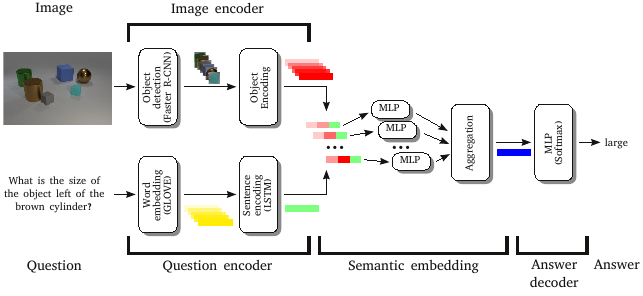}}
\caption{General architecture of the proposed system}
\label{fig:scheme_general}
\end{figure*}

The third direction results from the observation that different tasks might require totally different reasoning processes. For example in the "query attribute" task the system should focus its attention on a single aspect of the scene, whereas in "counting" it should perform a sequence of actions, focusing on the objects of interest one by one.
This observation led to the development of Neural Module Networks~\cite{andreas2015deep}, where semantic parsers (e.g. Stanford Parser) were used for question-driven dynamic assembly of the internal structure of the system, which is different for every type of question.
Most recent advances include program-induction~\cite{johnson2017inferring} consisting of a program generator and program executor, where the former is a sequence-to-sequence model (a pair of LSTMs, i.e. Long Short Term Memory recurrent neural nets~\cite{hochreiter1997long}) responsible for constructing of an internal representation of the reasoning process to be performed, whereas the latter is responsible for execution of the resulting program in order to produce the answer.

The last direction includes research on models using external knowledge bases (i.e. external sources of data) for answering the question.
One such solution is proposed in ~\cite{wu2016ask}, where the authors developed a framework that extracts image-related information from the DBpedia knowledge base, enabling it to answer a broad range of questions, often requiring knowledge that could not be inferred directly from the image.
The improvements achieved by results e.g. in the paper proposing the FVQA dataset~\cite{wang2017fvqa} strongly indicate that utilization of supporting-facts extracted from large-scale structured knowledge bases might be the key ingredient for solving the Visual Turing test. 
As there are other findings that also suggest the need for operation on high-level, abstract facts extracted from the image (e.g.~\cite{anderson2017bottom}), we decided to investigate that direction further. 


%% file: our_solution.tex
\section{Our solution}
\label{sec:our_solution}
Architecture of our system is presented in \fig{fig:scheme_general}. It extends the Encoder-Decoder~\cite{cho2014learning} architecture, which originally consisted of two RNNs, the first one used for encoding a sequence of input symbols into a fixed length representation, and the other for decoding that representation into another sequence of output symbols. 

In VQA there are two input channels with different modalities, thus we need two types of encoders and an intermediate module responsible for projection of both the visual and language representations into a common semantic embedding space.
Following that scheme, Xiao et al. \cite{xiao2017weakly} adopted the convolutional layers of the VGG network followed by an average pooling across the output features as the "visual" encoder and used a two-layer stacked LSTM network as the "language" encoder.
The authors assumed that the output of the language encoder is already in the semantic space and they used a two-layer-perceptron to project the output of the visual encoder into the same space.

Similarly, the Refined Ask Your Neurons architecture~\cite{malinowski2017ask} consisted of a visual encoder, question encoder, a multimodal embedding module combining both encodings into a joint space, and an answer decoder. The authors compared results achieved for several combinations of different question (e.g. BoW, CNN, GRU and LSTM) and visual (several classical architectures including AlexNet, GoogLeNet, VGG-19, ResNet-152) encoders.

\subsection{Image encoder}
\begin{figure}[htbp]
\centerline{\includegraphics[width=0.65\columnwidth]{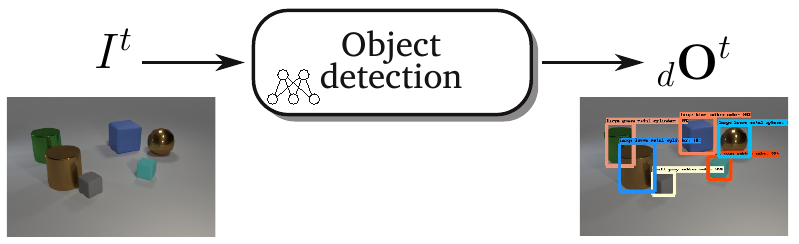}}
\caption{Dataflow diagram of object detection}
\label{fig:scheme_object_detection}
\end{figure}

In our system the image encoder realizes two major operations: object detection (OD) and object encoding.
The diagram for object detection is shown in \fig{fig:scheme_object_detection}.
In a given $t$-th image $I^t$ we detect a set of $N$ objects:
\begin{equation}
_d\textbf{O}^t = [_{d}o^t_0,\ _{d}o^t_1,\ \ldots,\ _{d}d^t_{N-1}  ],
\end{equation}
each defined as $_{d}o^t_n = < _{d}i^t_n,\ _{d}b^t_n > $, where $_{d}i^t_n$  denotes the class identifier (1 to 96) of the object and $_{d}b^t_n$ contain four parameters defining the object bounding box.
We decided to use Faster R-CNN~\cite{ren2015faster} with ResNet-101~\cite{he2016deep} pre-trained on COCO dataset, since it is one of the most reliable methods for object detection currently available.

\begin{figure}[htbp]
\centerline{\includegraphics[width=1.0\columnwidth]{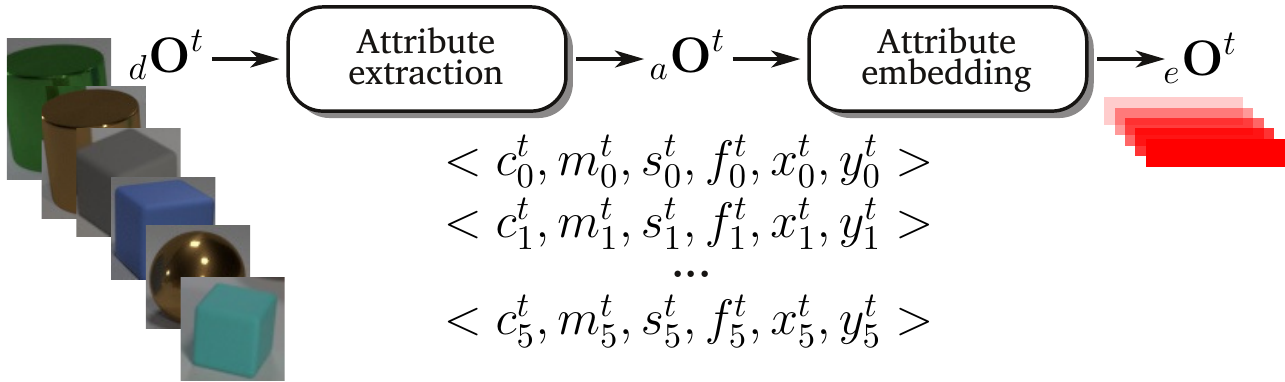}}
\caption{Dataflow diagram of object encoding}
\label{fig:scheme_object_decoding}
\end{figure}

In the next step (presented in \fig{fig:scheme_object_decoding}) we decode each object $_{d}o^t_n$ and retrieve set of its attributes:
\begin{equation}
_{a}o^t_n = < c^t_n,\  m^t_n,\ s^t_n,\ f^t_n,\ x^t_n,\ y^t_n >,
\label{eq:obj_attributes}
\end{equation}
where 
$c^t_n$ is the object color,
$m^t_n$ represents material it is made from,
$s^t_n$ indicates size,
$f^t_n$ represents its shape (form)
and $x^t_n$ and $y^t_n$ describe its position in the image.
Simple encodings of each of those attributes (we used one-hot in case of the former four and bucketing in the case of positional arguments, as explained in sec. 4.4) and concatenation of the resulting vectors form the encoded object description $_{e}o^t_n$.

\subsection{Question encoder}
\begin{figure}[htbp]
\centerline{\includegraphics[width=0.95\columnwidth]{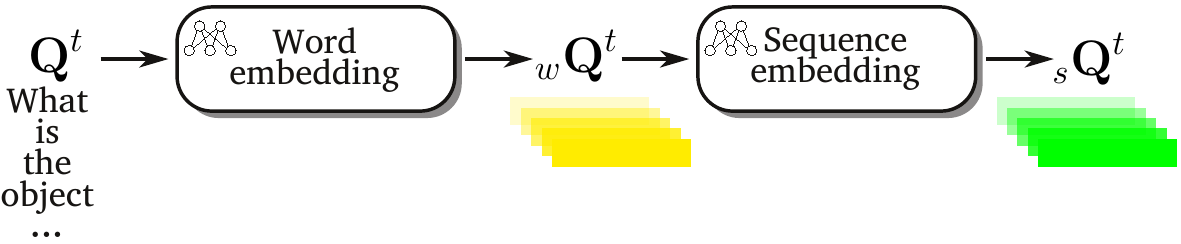}}
\caption{Dataflow diagram of question encoding}
\label{fig:scheme_question_encoding}
\end{figure}

The diagram for question processing is shown in \fig{fig:scheme_question_encoding}.
We start with the question consisting of several words:
\begin{equation}
\textbf{Q}^t = [q^t_0,\ q^t_1,\ \ldots,\ q^t_{W-1}  ],
\end{equation}
where $W$ denotes the number of words constituting a given question.
Next, we use the GloVe word embedding model~\cite{pennington2014glove} to encode question words: 
$_e\textbf{Q}^t = [_eq^t_0,\ _eq^t_1,\ \ldots,\ _eq^t_{W-1}  ] $.
Finally, we pass the encoded words one by one as inputs to the LSTM~\cite{hochreiter1997long} to produce a list of encoded output:
\begin{equation}
_{s}\textbf{Q}^t = [_{s}q^t_0,\ _{s}q^t_1,\ \ldots,\ _{s}q^t_{W-1}  ].
\end{equation}

\subsection{Semantic embedding}
There are many ways for combining image and question features~\cite{teney2017visual}, such as concatenation, element-wise product or bilinear operation.
In our solution, similar to the relational network module reported in ~\cite{santoro2017simple}, we form a set of feature vectors by concatenating pairs of vectors representing two objects with the encoded question, i.e. the last output of LSTM ($_{s}q^t_{W-1}$).
This approach is somehow similar to object-object-relation triplets used in~\cite{dai2017detecting}. 

\begin{figure}[htbp]
\centerline{\includegraphics[width=\columnwidth]{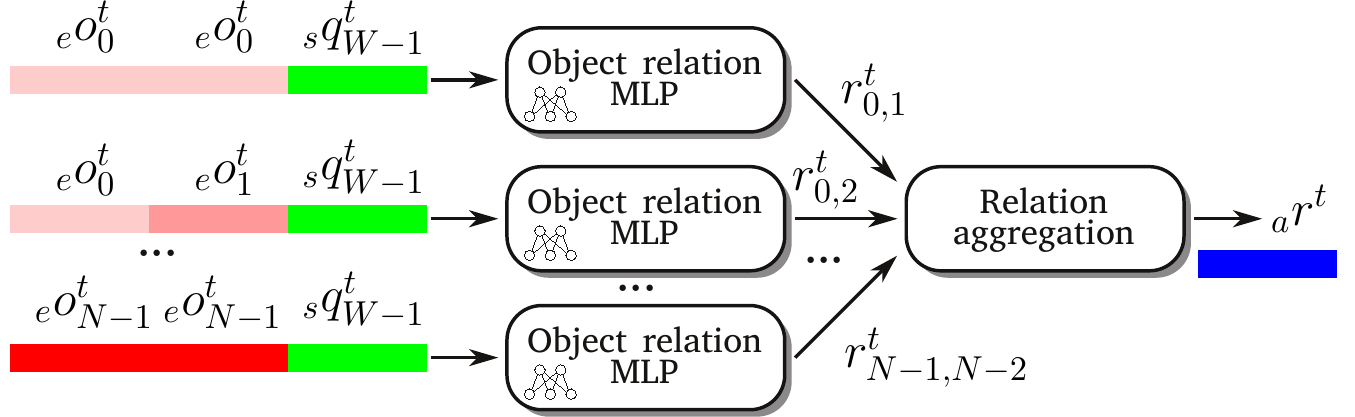}}
\caption{Dataflow diagram of semantic embedding}
\label{fig:scheme_semantic_embedding}
\end{figure}

Then, individual triplets are passed through a four layer MLP (Multi-Layer Perceptron), each with 512 units and ReLU non-linearities, which results in a vector of relations:
\begin{equation}
\textbf{R}^t = [r^t_{0,1},\ r^t_{0,1},\ \ldots,\ r^t_{N-1,N-1}].
\end{equation}

Those relations are further aggregated into $_ar^t$. 
We investigated several aggregation methods, but simple summation appeared to give the best results -- please refer to sec.~\ref{sec:results_clevr_baselines}.

\subsection{Answer decoder}
\begin{figure}[htbp]
\centerline{\includegraphics[width=0.8\columnwidth]{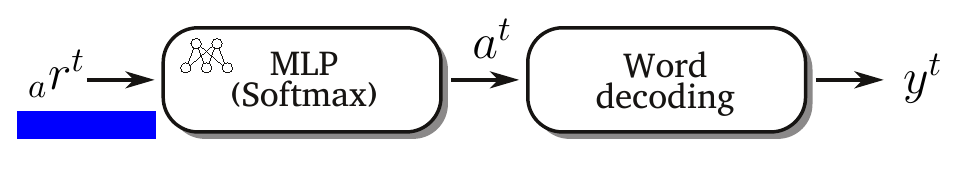}}
\caption{Dataflow diagram of answer decoding}
\label{fig:scheme_answer_decoding}
\end{figure}

Finally, we cast the answering problem as a classification task and pass the aggregated relation  $_ar^t$ through three MLP layers consisting of 512, 1024, and 29 units. A 2\% dropout layer was added before the last MLP layer. ReLU non-linearities were added in each layer except the last one, where we applied the softmax function. The result from the softmax was finally decoded using the word dictionary used for initial word embeddings in Question encoder (\fig{fig:scheme_answer_decoding}).

%% file: results.tex
\section{Analysis of the results}
\label{sec:analysis_results}
\subsection{Question encoding}
We have pretrained the GloVe word embedding model on the Google News corpus~\cite{mikolov2013efficient} (3 billion running words) word vector model and then generated embeddings for the dictionary formed on the basis of 93 words present in CLEVR questions and answers.

\begin{figure}[!t]
\centerline{\includegraphics[width=0.8\columnwidth]{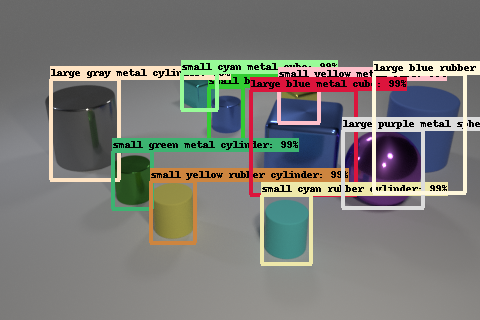}}
\caption{Exemplary CLEVR scene with objects detected with our trained object detector}
\label{fig:od_complex_scene_ok}
\end{figure}

\subsection{Image encoding}
\label{sec:image_encoding}
Utilization of object detector required its prior training on images with ground truth bounding boxes and object classes.
Since scene descriptions in the CLEVR dataset do not provide bounding boxes, we wrote a program for their generation. 
We calculated the  bounding boxes for every object present in the scene relying on the associated metadata from the scene description, i.e. using object position in the image (2d pixel coordinates and rotation), its position in the Cartesian space (3d coordinates) along with its size and shape.
In addition, we have assigned each object a class identifier taking into account the possible combinations of object attributes as defined in~\eqref{eq:obj_attributes}, which resulted in 96 unique classes.

An exemplary scene is presented in \fig{fig:od_complex_scene_ok}.
We compared the reconstructed scene from the OD and the ground truth scene from the scene description to check how accurately the OD network is predicting the scenes.
We were able to get a precision of 0.99 and a recall of 0.99, thus we find the results satisfactory.

\begin{figure}[b!]
\centerline{\includegraphics[width=\columnwidth]{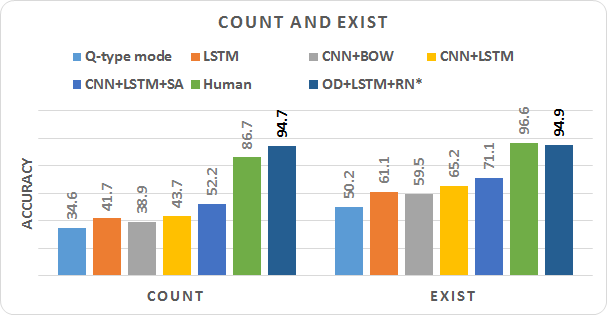}}
\caption{The comparison of accuracy of our solution in Count and Exist tasks with the CLEVR baselines}
\label{fig:count_exist}
\end{figure}

\begin{figure}[!t]
\centerline{\includegraphics[width=\columnwidth]{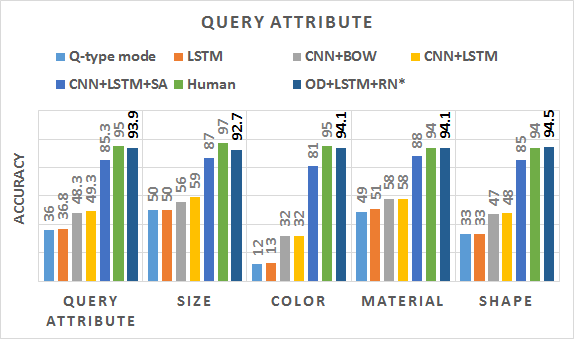}}
\caption{The comparison of accuracy of our solution with the CLEVR baselines in Query Attribute task}
\label{fig:query_attribute}
\end{figure}

\begin{figure}[!t]
\centerline{\includegraphics[width=\columnwidth]{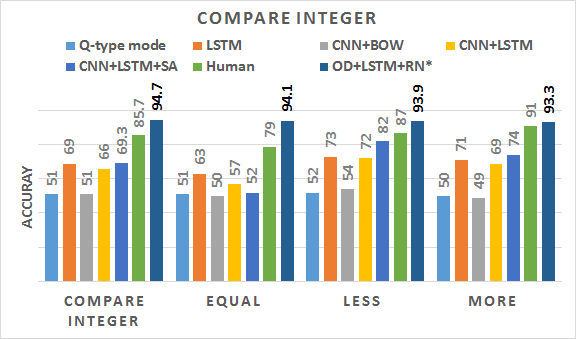}}
\caption{The comparison of accuracy of our solution with the CLEVR baselines in Compare Integer task}
\label{fig:compare_numbers}
\end{figure}

\begin{figure}[!t]
\centerline{\includegraphics[width=\columnwidth]{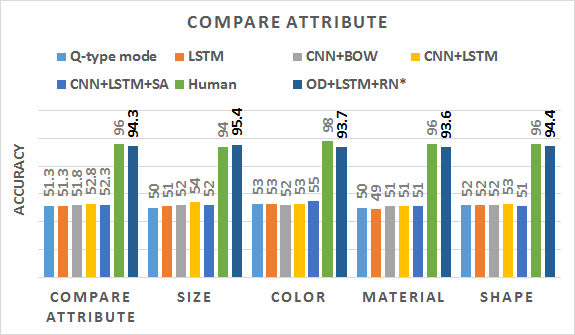}}
\caption{The comparison of accuracy of our solution with the CLEVR baselines in Compare Attribute task}
\label{fig:compare_attribute}
\end{figure}

\begin{figure*}[t!]
    \centering
    \begin{subfigure}[b]{0.33\textwidth}
        \centering
        \includegraphics[width=\textwidth]{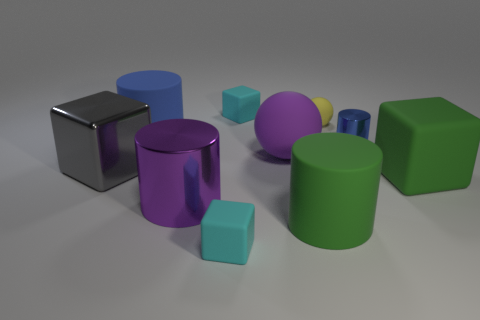}
        \caption{\textbf{Question:} what number of objects are small cyan objects that are in front of the gray object or big things that are in front of the big rubber block ?
\newline\textbf{Predicted Answer:} 3
\newline\textbf{Ground Truth:} 3}
    \end{subfigure}%
    ~ 
    \begin{subfigure}[b]{0.33\textwidth}
        \centering
        \includegraphics[width=\textwidth]{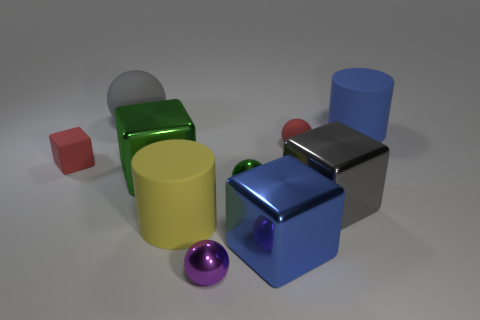}
        \caption{\textbf{Question:} how many balls are either big blue matte objects or small red things ?
\newline\newline
\newline\textbf{Predicted Answer:} 1
\newline\textbf{Ground Truth:} 1}
	\label{fig:counting_occluded_example1}
    \end{subfigure}%
    ~ 
    \begin{subfigure}[b]{0.33\textwidth}
        \centering
        \includegraphics[width=\textwidth]{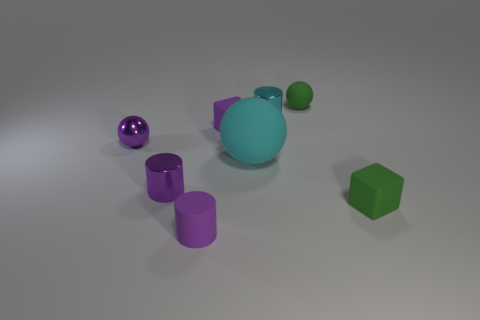}
        \caption{\textbf{Question:} what number of things are either small shiny objects that are behind the purple metallic cylinder or tiny purple metallic balls ?
\newline\textbf{Predicted Answer:} 2
\newline\textbf{Ground Truth:} 2}
	\label{fig:counting_occluded_example2}
    \end{subfigure}%
    
    \caption{Exemplary result on counting with heavily occluded object}
	\label{fig:counting_examples}
\end{figure*}

\subsection{Comparison with CLEVR baselines}
\label{sec:results_clevr_baselines}
We have reproduced the majority of baselines from the original CLEVR paper~\cite{johnson2017clevr}, achieving similar results.
In our implementations we used the TensorFlow framework~\cite{abadi2016tensorflow} and the associated object detection API~\cite{huang2016speed}.
As CLEVR does not provide ground truth for the test set, we are reporting the results achieved on the validation set.
During training we have used a learning rate of $1.0\*e^{-4}$.
In the last step of Semantic embedding we have tried several aggregation methods, including simple concatenation of activations of Object relation MLPs, using mean values of those activations and their sum.
The experiments have shown that summation of those activations gives the best results (accuracy of 94.6\% in comparison to e.g. 92.9\% in the case of using mean value).
The simple explanation is that summation ensures the invariance with respect to the order of objects~\cite{santoro2017simple}.

\begin{figure}[!b]
\centerline{\includegraphics[width=\columnwidth]{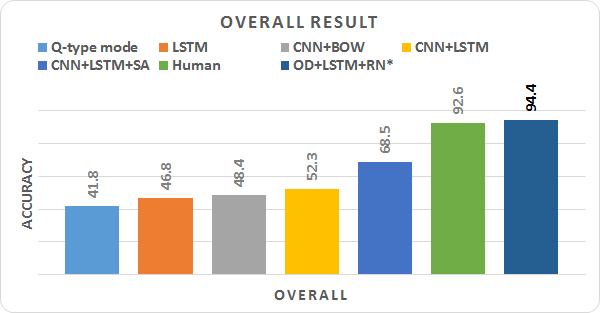}}
\caption{The comparison of overall accuracy of our solution with CLEVR baselines}
\label{fig:over_all}
\end{figure}

We have analyzed the results using the same task-oriented criteria as the original paper did.
The accuracy on "exist" and "count" tasks is presented in \fig{fig:count_exist}, where 
the existence questions ask whether a certain type of object is present, while the count questions ask for the number of objects fulfilling some conditions.
The Query Attribute task contains questions asking about an attribute of a particular object (\fig{fig:query_attribute}).  
The Compare Integer questions ask which of two object sets fulfilling given conditions is larger (\fig{fig:compare_numbers}). 
Results for Attribute Comparison task, where questions ask whether two objects have the same value for a given attribute, are presented in \fig{fig:compare_attribute}.

Finally, we present the comparison of the overall accuracy of our solution against all the CLEVR baselines in \fig{fig:over_all}.
Our solution has shown major improvement over all the baselines, including the human accuracy collected using Mechanical Turk.

\begin{figure}[!b]
\centerline{\includegraphics[width=\columnwidth]{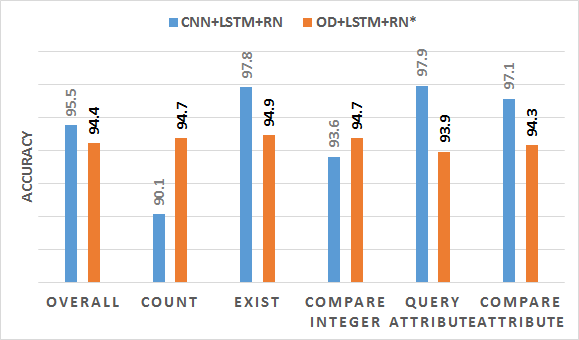}}
\caption{Comparison of our approach with results reported in the paper on Relational Networks}
\label{fig:compare_with_rn}
\end{figure}

\begin{figure*}[!t]
\centerline{\includegraphics[width=\textwidth]{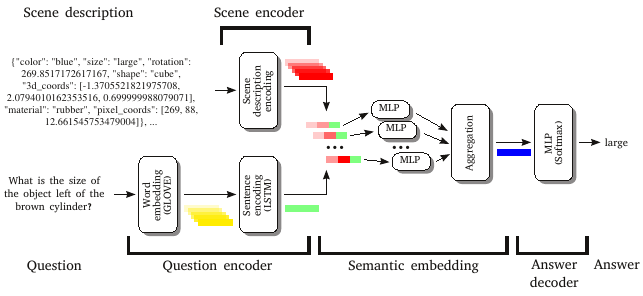}}
\caption{The architecture of the system with scene description as input used for comparison with the \textit{CLEVR with state description} baseline from~\cite{santoro2017simple}}
\label{fig:scheme_general_scene_description}
\end{figure*}

\subsection{Comparison with results from RN}
\label{sec:results_rn_baselines}
In \fig{fig:compare_with_rn} we present the comparison of our results (shown in orange) with the results reported in~\cite{santoro2017simple} (shown in blue).
As one might notice, our overall accuracy is slightly, by around 1\%, lower.
However, interestingly our solution seems to give more consistent results for different tasks and achieves better results in 2 out of 5 tasks.
In particular, we achieved much better accuracy (by almost 5\%) in the counting task, which is currently considered to be one of the hardest tasks in VQA (e.g.~\cite{trott2017interpretable}).
In order to realize the counting task, the system is supposed to perform several quite different operations, i.e. understanding what type of objects to focus on, finding those objects in the image and finally counting the instances.
Our results indicate that a prior object detection naturally facilitates counting, which we treat as support for our hypothesis that operation on more abstract facts facilitates reasoning in general.
In \fig{fig:counting_examples} we present a few hard cases, where our system was able to provide the correct answer.
In particular, it managed to properly answer questions about the scenes presented in \fig{fig:counting_occluded_example1}  and \fig{fig:counting_occluded_example2} despite the heavy occlusions of the objects of interest, which in~\cite{santoro2017simple} was suggested to be the main reason for failures in CLEVR.

We further investigated why our solution does not improve in the other three tasks.
After validating the accuracy of object detection module (as reported in \sec{sec:image_encoding}) we decided to reproduce the experiments from~\cite{santoro2017simple} in which the Image encoder was replaced by the information retrieved straight from the scene description.
The architecture of the resulting system is presented in \fig{fig:scheme_general_scene_description}.
Despite using the same learning hyperparameters for relational MLPs we achieved the overall accuracy of 94.5\%, which is almost 2\% worse from the accuracy of 96.4\% reported for that setting in the paper~\cite{santoro2017simple}.
However, as the authors did not report how they encoded the scene, we decided to further investigate that research direction.
As the object attributes form small dictionaries, we assumed that simple one-hot encoding is sufficient.
However, the object position expressed in pixel coordinates in the images ($x^t_n$ and $y^t_n$) might be encoded in several different ways, which in turn might influence the final accuracy. 
Therefore, we have performed several experiments with different methods of encoding the object positions, briefly explained below.

\subsubsection{One-hot encoding}
In this approach we simply converted the object $x^t_n$ and  $y^t_n$ pixel coordinates into two separate vectors using one-hot encoding and concatenated them into one vector of length 800 (as we have 480 possible values for $x^t_n$ and 320 for $y^t_n$). As can be seen from tab.~\ref{tab:scene_enconding} this method was not able to generalize as well as the others. 

\begin{table*}[htbp]
  \centering
	\begin{center}
	\begin{tabular}{|c|c||c|c|c|c|c|c|}\hline 
	\textbf{Encoding} & \textbf{Bucket} & \textbf{Overall} & \textbf{Count} & \textbf{Exist} & \textbf{Compare} & \textbf{Query} & 	\textbf{Compare} \\ 
 	& \textbf{size} & & & & \textbf{numbers} & \textbf{attribute} & \textbf{attribute} \\
	\hline 
	\hline 
    One hot& -- & 89.8 & 90.5 & 89.7 & 89.5 & 90.6 & 88.5 \\ 
	\hline 
    Bucketing& 15 & 93.9 & 93.9 & 93.7 & 93.6 & 94.6 & 93.5 \\ 
	\hline 
	Bucketing& 20 & \textbf{94.5} & 93.6 & 94.7 & 93.3 & 95.2 & 94.4 \\ 
	\hline 
	Bucketing& 30 & 93.6 & 93.4 & 93.1 & 93.2 & 94.9 & 93.4 \\ 
    \hline 
    Enumeration & -- & 93.3 & 93.2 & 93.7 & 92.9 & 93.5 & 93.1 \\ 
	\hline 
	\hline 
    Results from~\cite{santoro2017simple} & -- & 96.4 & -- & -- & -- & -- & -- \\ 
	\hline 
	\end{tabular} 
	    \end{center}
    \caption{Results on different position encodings for solution with Image encoder replaced by the information retrieved straight from the scene description}
    \label{tab:scene_enconding}
\end{table*}

\subsubsection{Bucketing}
Here we grouped n consecutive pixels to buckets along the x and y axes separately, which resulted in a number of buckets for x and y. 
We did experiments on three different bucket sizes, i.e. 15, 20, and 30, which resulted in 32, 24, and 16 buckets for the x component and 22, 16, and 11 buckets for the y component respectively.
When given the object position coordinate $x^t_n$ belongs to a range of a given bucket we convert the bucket number to vector using one hot encoding. We perform the same for $y^t_n$ 
and concatenate both results.
It can be observed in tab.~\ref{tab:scene_enconding} that the solution generalizes better when the bucket size is chosen to be 20.

\subsubsection{Object enumeration}
In this method, instead of relying on the pixel coordinates we move to a more abstract representation and encoded the position of the object in relation to the positions of other objects in the scene. For x and y axes we form two separate lists, representing the order of the objects according that axis.
For instance, for three objects in a scene, the very top object out of the three will have a value of one and the bottom object will have a value of three.
Finally, we encoded such represented positions using one hot encoding and concatenated both encoded coordinates for every object.

Results of all the above mentioned experiments are presented in Tab.~\ref{tab:scene_enconding}.
The main finding is that our model was able to generalize well when we grouped twenty pixel values into one bucket.
Unfortunately, these experiments did not improve the accuracy to the level reported in ~\cite{santoro2017simple}.